\documentclass[runningheads]{llncs}
\usepackage[T1]{fontenc}
\usepackage{multirow}
\usepackage{graphicx}
\usepackage{graphicx,verbatim}
\usepackage{amsmath} 
\usepackage{array}
\usepackage{amssymb}
\usepackage{colortbl}
\usepackage{arydshln}

\newcommand{\bzeta}{\boldsymbol{\zeta}}

\begin{document}
\title{4D Triangle Splatting Reconstruction of Dynamic Endoscopic Scenes from Monocular Videos}
\titlerunning{4D Triangle Splatting for Endoscopy}
%
\author{Laura Salort-Benejam\inst{1}\orcidID{0009-0000-6864-0854} \and
Stamatia Giannarou\inst{2}\orcidID{0000-0002-8745-1343} \and
Antonio Agudo\inst{1}\orcidID{0000-0001-6845-4998}}
\authorrunning{L. Salort-Benejam et al.}
%
\institute{Institut de Robòtica i Informàtica Industrial, CSIC-UPC, Barcelona, Spain \email{\{laura.salort, antonio.agudo\}@upc.edu}\\
\and
Imperial College London, London, United Kingdom\\
\email{stamatia.giannarou@imperial.ac.uk}}

\maketitle              
\begin{abstract}

Robust 4D reconstruction of deformable endoscopic scenes from monocular videos remains challenging due to non-rigid tissue deformation, illumination changes, occlusions, and unknown camera trajectories. Reliable dynamic reconstruction of endoscopies could significantly enhance visualization, diagnostic accuracy and intraoperative guidance of minimally invasive procedures. We propose SurgicalTS, a self-supervised triangle-splatting framework for dynamic endoscopic scene reconstruction from static monocular cameras. The scene is represented by a canonical set of triangles jointly optimized with a deformation network. We introduce a quaternion-based deformation model, an improved color param-etrization, spatio-temporal priors, and a redefined initialization strategy that improves completeness under occlusions. Extensive evaluation on challenging in-vivo endoscopy videos from diverse procedures shows that SurgicalTS successfully models soft-tissue deformations and tool-tissue interactions over time, while consistently outperforming state-of-the-art methods. Our method also reduces training times while maintaining competitive geometric and photometric accuracy results, highlighting the effectiveness of quaternion-based deformation modelling with physical consistency for robust and efficient 4D reconstruction of endoscopic scenes.


\end{abstract}
\section{Introduction}

Minimally invasive endoscopic procedures are common practice for diagnosis and treatment across multiple medical specialties, such as gastrointestinal endoscopy, bronchoscopy, arthroscopy or laparoscopy. The use of endoscopes~\cite{YANG2024108546} enables direct visualization of the internal anatomy while reducing the physical trauma and long recovery times of open surgery. Despite its widespread use, most endoscopes are equipped with monocular RGB cameras, prioritizing compactness and versatility over obtaining direct depth information from RGB-D sensors or stereo cameras. 

Recovering 3D information from visual cues in monocular endoscopy remains a major challenge, as the problem is inherently ill-posed and faces more difficulties than natural-scene 3D reconstruction such as non-rigid deformations, specular reflections from the light source, illumination changes, motion blur, frequent occlusion caused by surgical instruments, limited viewpoints or low-texture regions. Overcoming these obstacles to obtain a robust method for 3D reconstruction would be highly beneficial for clinical practice by enabling improved visualization of complex anatomical regions, assisting diagnosis or treatment decisions, facilitating intraoperative guidance with temporally consistent patient-specific models, and for postoperative monitoring.

Traditional approaches, including rigid and non-rigid Structure-from-Motion (SfM)~\cite{agudoICPR2020,DSfMcvpr25,schoenberger2016sfm}, shape-from-template~\cite{ROBUSfT_imavis24}, photometric stereo~\cite{photometric-stereo}, and shape-from-shading~\cite{SmeetsM_cvpr25,LightDepth}, depend on explicit feature correspondences, restrictive deformation priors or controlled lighting assumptions that rarely hold in endoscopy settings. The introduction of Neural Radiance Fields (NeRF)~\cite{original-NERF} redefined 3D reconstruction and novel-view synthesis by learning a differentiable implicit rigid scene representation directly from visual information, but it relies on known camera poses for initialization and has a high computational cost, motivating extensions for accelerated training times~\cite{fridovich2022plenoxels,instant-npg15}, camera pose estimation~\cite{SCNeRF2021,bundlesdfwen2023}, and deformable environments~\cite{4DPV,nerfies,d-nerf}. Several adaptations to deformable endoscopy settings~\cite{salortISBI26,endoNERF,lerplane,endoSURF} have been proposed, often assuming fixed cameras to make the reconstruction problem more manageable, which is a reasonable assumption for many procedures where the endoscope is static to allow the physician more precise assessment and manipulation of the anatomy. Nonetheless, some are also dependent on stereo inputs, limiting their applicability to a smaller number of procedures where this information is available.

Explicit methods such as 3D Gaussian Splatting (3DGS)~\cite{original-gs} emerged as a faster and high quality alternative, yet the original formulation assumes rigid scenes and relies on SfM-based initialization. Subsequent dynamic variants~\cite{per-gaussian,spacetime-gs,gaussian-flow,dynamic3dgs,4DGS,motiongs} incorporate deformation networks to model time-varying geometry but remain constrained by these limitations. Related methods for deformable endoscopy reconstruction~\cite{endo4dgs,endogaussian,surgicalgaussian,deform3dgs,endogs} improve efficiency and reconstruction quality but inherit the assumptions of previous NeRF-based endoscopic methods. Endoscopic SLAM approaches~\cite{pancakes,endogslam} have also been proposed mainly addressing camera pose estimation rather than jointly modelling tissue deformation. More recently, triangle splatting~\cite{trianglesplatting26} introduced a differentiable rendering framework based on unstructured, disconnected triangle primitives. The use of triangles instead of Gaussians as primitives allows for faster rendering, given that modern hardware pipelines are optimized to process and render triangles at high speeds. Additionally, Gaussian primitives have unbounded support while triangles are bounded, making the rendering equation faster to compute as they contribute to a smaller number of pixels and this also makes it easier to optimize their positions and appearance to the given geometry. This method was later extended by Endo-4DTS~\cite{salort-benejam2026endodts} to deformable endoscopic scenes without requiring SfM calibration, pre-computed templates, stereo depth information or feature-based approaches for initialization. Although Endo-4DTS obtained strong results in terms of visual metrics, we observed that it failed in jointly modelling the geometry and appearance of the scene, resulting in considerable discontinuities in the depth maps. 

In this work, we propose SurgicalTS a novel physically-aware triangle-splatting framework for 3D reconstruction of deformable endoscopic scenes from monocular videos. Our main contributions are: 1) a refined initialization strategy to enhance the completeness of the triangle representation in the presence of occlusions; 2) new color and motion parametrizations learned through a deformation network; 3) spatio-temporal physical priors to improve the robustness and reconstruction accuracy of our method. To the best of our knowledge, SurgicalTS is the first triangle splatting approach to propose these contributions for deformable monocular endoscopy reconstruction. Experimental results demonstrate competitive or superior performance to state-of-the-art methods in both visual and geometric metrics, while improving convergence and rendering efficiency.
\section{Methodology}

In this work, we present a new deformable reconstruction method for endoscopic scenes, based on triangle splatting~\cite{trianglesplatting26,salort-benejam2026endodts}. In the seminal work~\cite{trianglesplatting26}, a conventional rigid scene is represented as a set of learnable 3D triangle primitives, each defined by three vertices $\textbf{v}_i \in \mathbb{R}^3$, a color $\textbf{c}$, an opacity $o$, and a smoothness parameter $\sigma$. The triangles are projected onto the image plane using a standard pinhole camera model as $\textbf{p}_i=\textbf{K}(\textbf{R}_{cam}\textbf{v}_i+\textbf{t}_{cam})$, where $\textbf{p}_i\in \mathbb{R}^3$ represents the homogeneous coordinates in image space of the projected triangle vertex, $\textbf{R}_{cam}\in \mathbb{R}^{3\times3}$ and $\textbf{t}_{cam}\in \mathbb{R}^{3}$ are the rotation and translation components, respectively, that determine the camera pose, and $\textbf{K}\in \mathbb{R}^{3\times3}$ is the intrinsic camera matrix. The final color $\bzeta$ at pixel $\textbf{p}$ is obtained by following the rendering equation in~\cite{salort-benejam2026endodts}.

\begin{figure}[h] 
    \centering
    \includegraphics[width=1\linewidth]{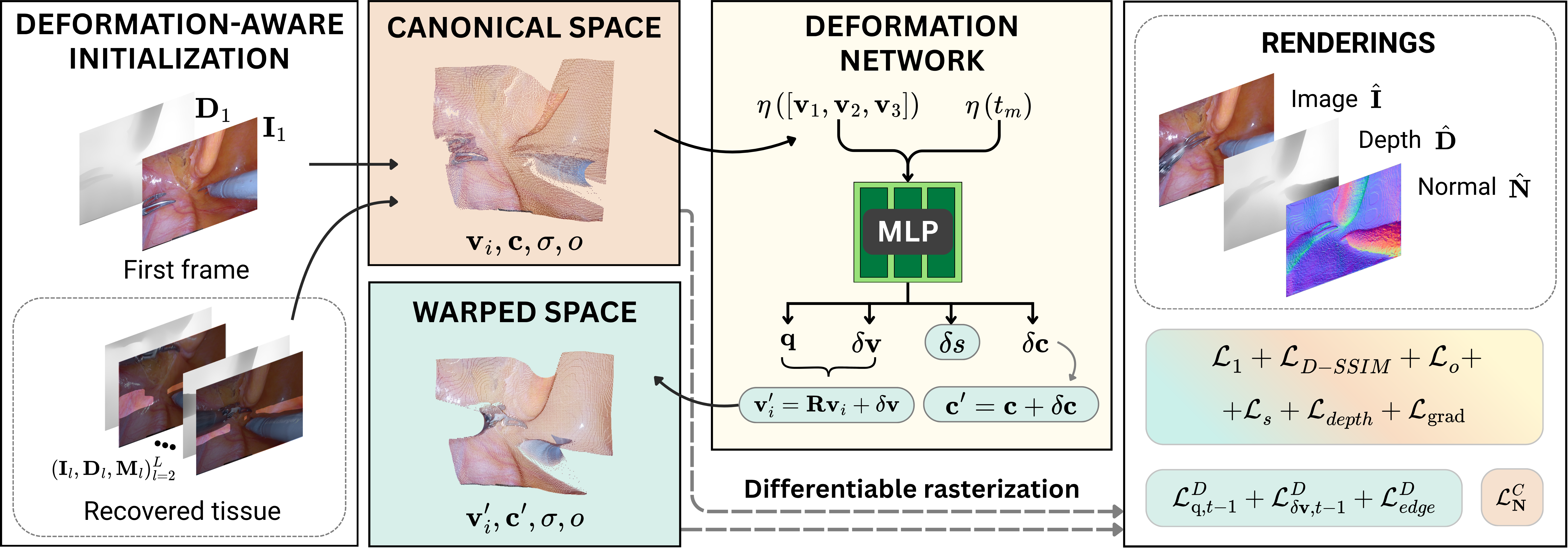}
    \centering
\caption{\textbf{Overview of our SurgicalTS} pipeline for triangle splatting 4D reconstruction of dynamic monocular endoscopy videos. We optimize a canonical space with a deformation network to model the scene geometry and appearance over time.} \label{pipeline}
\end{figure}

We adopt this representation in a canonical space, where a static set of 3D triangles is jointly optimized with a deformation network that maps the canonical primitives to a warped space at each time step by predicting per-triangle rotations, translation, scale and color adjustments, as is shown in Fig.~\ref{pipeline}.

Given a monocular video composed of $L$ images, the inputs to our model are defined as $\bigl\{(\textbf{I}_l,\textbf{D}_l, \textbf{M}_l, t_l)_{l=1}^L, \textbf{P},f\bigl\}$, where for each $l$-th frame we consider the RGB image $\textbf{I}_l\in\mathbb{R}^{H\times W\times3}$ of height $H$ and width $W$, a monocular depth map  $\textbf{D}_l \in \mathbb{R}^{H \times W}$ extracted from $\textbf{I}_l$ by video depth anything~\cite{video_depth_anything}, a binary segmentation of the surgical tools $\textbf{M}_l \in \mathbb{R}^{H \times W}$ , the corresponding time index $t_l = l/L \in [0,1]$, the extrinsic camera matrix $\textbf{P} \in \mathbb{R}^{4\times4}$ representing a static camera pose, and the focal length of the camera $f$.

\subsection{Deformation-aware initialization}\label{sec:init}

Splatting-based methods usually initialize primitives using an SfM point cloud~\cite{trianglesplatting26,original-gs}. However, in endoscopic settings SfM becomes unreliable, and prior works typically rely on depth information for initialization, and tend to exclude the surgical instruments from the scene~\cite{endogaussian,surgicalgaussian}. In contrast, we explicitly include the tools in our model. From a clinical standpoint, preserving the instrument geometry is preferable to allowing the model to synthesize tissue in regions that remain occluded throughout the video--an artifact that can easily misguide the physician during its assessment. For instance, this could be particularly important for applications involving measurements, where hallucinated tissue could directly lead to erroneous estimations of tumor or resection volumes. Furthermore, instead of initializing solely from the information in the first frame as in~\cite{endo4dgs,salort-benejam2026endodts}, we aggregate depth and color information from subsequent frames to recover regions occluded in earlier views, enabling a more complete initialization. In particular, a $l$-th point cloud can be inferred by means of:
\begin{equation}
    \textbf{C}_l(\textbf{p})=\textbf{K}_e^{-1}\textbf{K}^{-1}\textbf{D}_l(\textbf{p})(\textbf{I}_l \odot (\textbf{1}- \textbf{M}_l))(\textbf{p}),
\end{equation} 
where $\odot$ denotes a Hadamard product, $\textbf{1}$ is a matrix of ones and $\textbf{K}_e$ and $\textbf{K}$ represent the extrinsic and intrinsic camera matrices, respectively. The first point cloud $\textbf{C}_1(\textbf{p})$ can be extended as:
\begin{equation}
\textbf{C} = \textbf{C}_1 \;\; \cup_{l=2}^L  (\textbf{C}_l \odot \textbf{B}_l) \;,
\end{equation}
where $\cup$ denotes a disjunction operator to include more information with respect to $\textbf{C}_1$. In the previous equation, we introduce a time-varying mask defined as:
\begin{equation}
    \textbf{B}_l= \mathbb{I} (|\textbf{D}_1-\textbf{D}_l|> \varphi) \cup (\textbf{M}_1 \cap (\textbf{1}-\textbf{M}_l)),
\end{equation}
where $\mathbb{I}$ is an indicator function and $\varphi$ a scalar to measure the depth variance. Thanks to that dynamic mask, we detect occluded tissues or significant depth variations to extract additional points.

\subsection{Deformation Network}\label{sec:deformation}
The deformation network $S_\Upsilon$ consists of an 8-layer multi-layer perceptron~\cite{surgicalgaussian,deformable3dgs}. Its inputs are the triangle vertices in the canonical space $\textbf{v}_i$ and the time step $t_l$ of the current frame, and positional encoding $\eta(\cdot)$ is applied to both inputs following~\cite{original-NERF}. 
For each triangle, the network predicts a rotation represented as a unit quaternion $\textbf{q}$, a translation offset $\delta\textbf{v}\in \mathbb{R}^3$, a scaling offset $\delta s\in \mathbb{R}$, and a color offset $\delta \textbf{c}\in \mathbb{R}^3$ as:
\begin{equation}
    \bigl(\textbf{q},\delta\textbf{v},\delta s,\delta\textbf{c}\bigl)  = S_\Upsilon\Bigl(\eta \bigl([\textbf{v}_1,\textbf{v}_2,\textbf{v}_3]\bigl),\eta(t_l)\Bigl).
\end{equation}

We adopt quaternions for the rotation representation instead of the Euler angles used in~\cite{salort-benejam2026endodts}, avoiding gimbal lock and improving numerical stability during optimization. The quaternions are normalized to unit length and converted to a rotation matrix $\textbf{R}\in SO(3)$, which we then apply to each vertex of the canonical triangles to obtain their warped position as $\textbf{v}_i'= \textbf{R}\textbf{v}_i+\delta\textbf{v}$.
Finally, we also model the appearance change in the canonical triangle differently, by defining it as an offset added to the canonical color instead of a completely new color estimation, given by $\textbf{c}'=\textbf{c}+\delta\textbf{c}$.

\subsection{Optimization}\label{sec:optim}
In our training function, the canonical triangle representation is jointly optimized with the deformation network through a combination of photometric, geometric and spatio-temporal regularization terms. Our loss function can be written as:
\begin{align}\label{eq:total-loss}
    \mathcal{L} =\; 
    &\alpha_0((1-\alpha_1)\mathcal{L}_1 
    + \alpha_1 \mathcal{L}_{D\text{-}SSIM})
    + \alpha_2 \mathcal{L}_o
    + \alpha_3 \mathcal{L}_s 
    + \alpha_4 \mathcal{L}_{\mathrm{depth}} \\
    &+ \alpha_5 \mathcal{L}_{\mathrm{grad}}
    + \alpha_6 \mathcal{L}^C_{\mathbf{N}}
    + \alpha_{7} \mathcal{L}_{\mathbf{q},t-1}^D
    + \alpha_{8} \mathcal{L}_{\delta\mathbf{v},t-1}^D 
    + \alpha_{9} \mathcal{L}_{\mathrm{edge}}^D ,\nonumber
\end{align}

where $\alpha_{0-9}$ denote weighting factors. Loss terms without any superscript are applied to both the canonical and warped spaces, and those with a superscript $^C$ or $^D$ are only applied to the canonical and warped spaces, respectively. 
First, color consistency between rendered and input images is enforced, following rendering approaches, by means of a photometric loss through a combination of an $l_1$-norm and a Differential-Structural Similarity Index Measure (SSIM) loss as  $\mathcal{L}_{D-SSIM}=1-SSIM(\textbf{x},\textbf{y})$. To prevent triangles from becoming fully transparent or saturated we use the opacity regularization from~\cite{mcmc} by means of the loss $\mathcal{L}_o=\frac{1}{N_T}\sum_{n=1}^{N_T}|o_n|$, where $N_T$ is the total number of triangles. To avoid excessively small or degenerate triangles we penalize these primitives using the term $\mathcal{L}_s=2\|(\textbf{v}_1-\textbf{v}_0)\times(\textbf{v}_2-\textbf{v}_0)\|_2^{-1}$~\cite{trianglesplatting26,salort-benejam2026endodts}. 

We also supervise the geometry of the scene $\hat{\textbf{D}}(\textbf{p})$ using the {\em pseudo-ground truth depth maps} ${\textbf{D}_l}(\textbf{p})$, encouraging triangles to remain close to the pseudo-surface through an $l_1$-norm between the input and rendered depth maps as:
\begin{equation}
    \mathcal{L}_{depth}=\| \hat{\textbf{D}}(\textbf{p}) -\textbf{D}(\textbf{p})\|_1.
\end{equation}

The previous term is complemented by including a gradient regularization to encourage the edges or discontinuities in the rendered depth to be similar to the input one. Inspired by~\cite{salortISBI26}, we propose the $l_1$-norm between the spatial gradients of the input and rendered depth maps as:
\begin{equation}
    \mathcal{L}_{grad}=\| \nabla_x \hat{\textbf{D}}(\textbf{p}) -  \nabla_x \textbf{D}(\textbf{p})\|_1 + \| \nabla_y\hat{\textbf{D}}(\textbf{p}) - \nabla_y \textbf{D}(\textbf{p}) \|_1.
\end{equation}

To encourage the rendered surface normal $\textbf{n}$ to be aligned with the pseudo-ground truth normals $\textbf{N}$, increasing so the robustness of the method, we apply the following term~\cite{salort-benejam2026endodts} to the canonical triangles:
\begin{equation}
     \mathcal{L}_\textbf{N}^C=1-\textbf{n}(\textbf{p})^\top\textbf{N}(\textbf{p}).
\end{equation}

Finally, we propose to use spatio-temporal physical priors to provide better consistency in the deformation. To this end, we impose temporal smoothness priors in both rotations and translations of the triangles by means of first-order filters. Moreover, we introduce a pairwise distance regularization that penalizes variations between the triangles' edges $\mathcal{E}= \{ (i,j)\; | \; 1\leq i < j \leq 3\}$ in the canonical $\textbf{v}$ and warped $\textbf{v}'$ states. These terms are written as:
\begin{align}
    \mathcal{L}_{\text{q,t-1}}^D &= \|\boldsymbol{\textbf{q}}_t-\boldsymbol{\textbf{q}}_{t-1}\|_2 , \\
        \mathcal{L}_{\delta \textbf{v},t-1}^D &= \|\boldsymbol{\delta \textbf{v}}_t-\boldsymbol{\delta \textbf{v}}_{t-1}\|_2 ,\\
            \mathcal{L}_{\text{edge}}^{D} &=
    \frac{1}{N_T}\sum_{n=1}^{N_T}\sum_{(i,j) \in \mathcal{E}}
    \Big| \, \| \mathbf{v}_i - \mathbf{v}_j \|_2 - \| \mathbf{v}'_{i} - \mathbf{v}'_{j} \|_2  \Big| .
\end{align}

\section{Experimental Evaluation}
We now present our experimental results on several challenging datasets. In particular, we use the Endo-NeRF~\cite{endoNERF} and NeRFscopy~\cite{salortISBI26} datasets, where up to six videos with different tissues, deformations and interventions are considered, and we follow previous works in using a 7:1 split for training and test frames. For further details, we refer the reader to these papers. To make a fair comparison with Endo-4DTS~\cite{salort-benejam2026endodts}, the only previous method based on triangle splatting, we exploit the same densification and pruning strategies as well as the number of iterations. The weights in Eq.~\eqref{eq:total-loss} were empirically set as: $\alpha_1=0.2,\alpha_2=0.0055,\alpha_3=1e^{-8},\alpha_{4,5}=1.0,\alpha_6=0.01,\alpha_{0,7,8}=10$ and $\alpha_9=0.1$.

Regarding quantitative evaluation, for photometric quality we provide Peak Signal-to-Noise Ratio (PSNR)~\cite{psnr}, SSIM~\cite{ssim} and Learned Perceptual Image Patch Similarity (LPIPS)~\cite{lpips}. Mean Average Error (MAE) and Root Mean Squared Error (RMSE) are considered for depth evaluation~\cite{LightDepth}, using the depth maps ${\textbf{D}_l}$ obtained from video depth anything~\cite{video_depth_anything} as a pseudo-ground given that no 3D ground truth information is available for this type of sequences.

\begin{table}[h!]
\renewcommand{\arraystretch}{1.2}
    \centering
    \caption{\textbf{Quantitative results of the ablation study} on the {\em pulling} sequence. Best results in \textbf{bold}, second best \underline{underlined}.}\label{tab:ablation}
    \begin{tabular}{c|ccc|cc}
        & {PSNR}$\uparrow$   & {SSIM}$\uparrow$  & {LPIPS}$\downarrow$  & {MAE}$\downarrow$  & {RMSE}$\downarrow$  \\
        \hline
        SurgicalTS & \underline{42.339}&	\underline{0.977} &	\textbf{0.014}  & \underline{1.825} &\underline{6.703}  \\ 
        w/o $\mathcal{L}_{\text{edge}}^{D}$ & 42.062	 &0.977 &	0.015 & \textbf{1.818} &\textbf{ 6.607}  \\
        w/o $\mathcal{L}_{\text{grad}}^{D}$ & \textbf{42.371} &\textbf{	0.978} &	\underline{0.014}  &1.957 & 7.753\\        
        w/o $\mathcal{L}_{\text{grad}}^{C} $ & 42.238 &	0.977 &	0.015   &1.876 & 7.115\\
        \noalign{\vskip 1pt}
        \hdashline
        \noalign{\vskip 1pt}
                w/o $\mathcal{L}_{\text{depth}}^{D} $ & 43.183 &0.980 &	0.011 & 3.585  &10.482 \\ 
        w/o $\mathcal{L}_{\text{depth}}^{C} $& 42.767 &	0.979 &	0.012  & 7.128 & 13.199 \\
        \hline
    \end{tabular}
\end{table}

\newcolumntype{?}[1]{!{\vrule width #1}}
\begin{figure}[h!]
    \centering    
    \scriptsize{
    \begin{tabular}[C]{c|cccccc}
        Reference 
        & SurgicalTS 
        & w/o $\mathcal{L}_{\text{edge}}^{D}$
        & w/o $\mathcal{L}_{\text{grad}}^{D}$ 
        & w/o $\mathcal{L}_{\text{grad}}^{C} $
        & w/o $\mathcal{L}_{\text{depth}}^{D} $
        & w/o $\mathcal{L}_{\text{depth}}^{C} $ \\
        \includegraphics[clip, angle=0, height=0.1\linewidth]{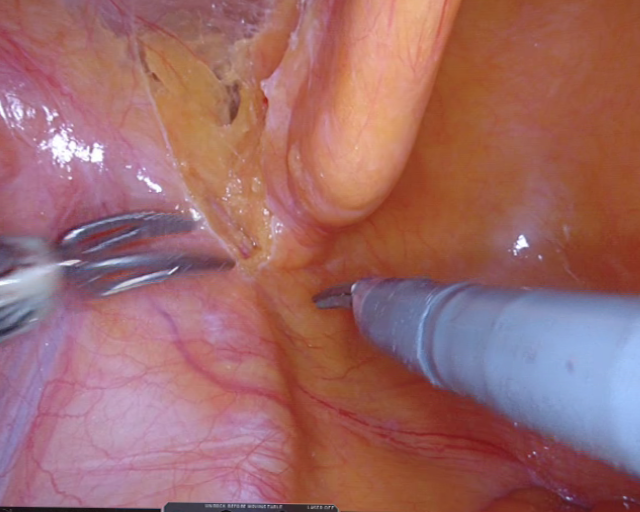}&
        \includegraphics[clip, angle=0, height=0.1\linewidth]{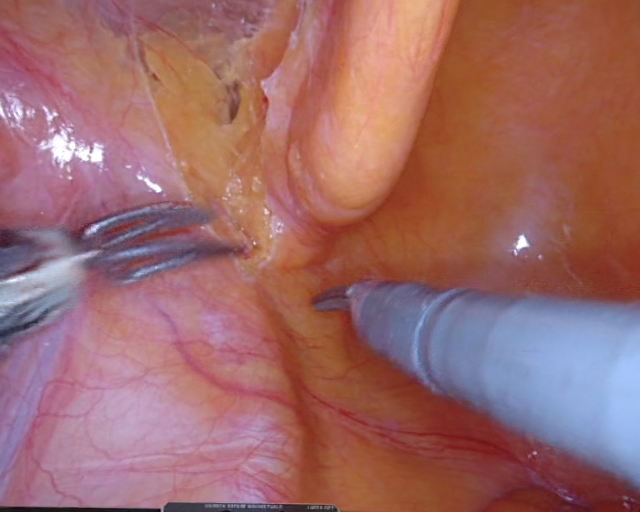}&
        \includegraphics[clip, angle=0, height=0.1\linewidth]{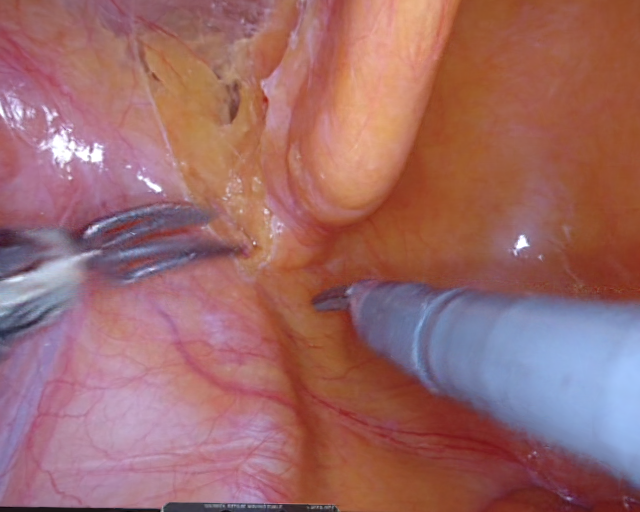}&
        \includegraphics[clip, angle=0, height=0.1\linewidth]{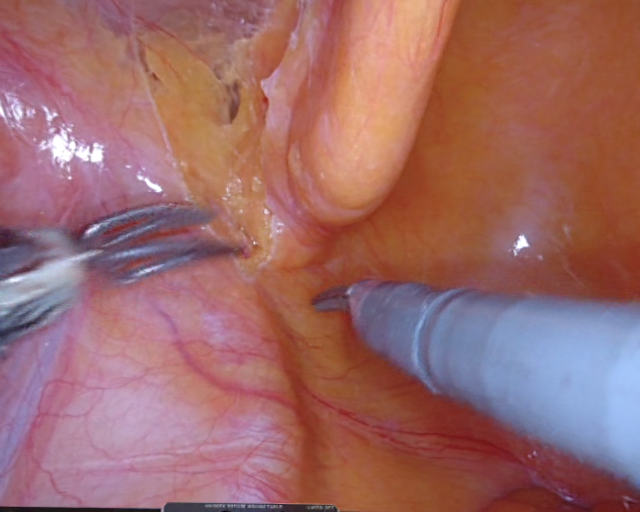}&
        \includegraphics[clip, angle=0, height=0.1\linewidth]{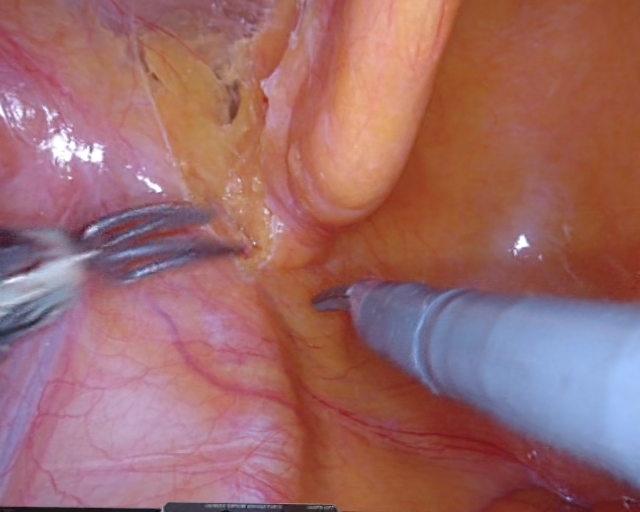}&
        \includegraphics[clip, angle=0, height=0.1\linewidth]{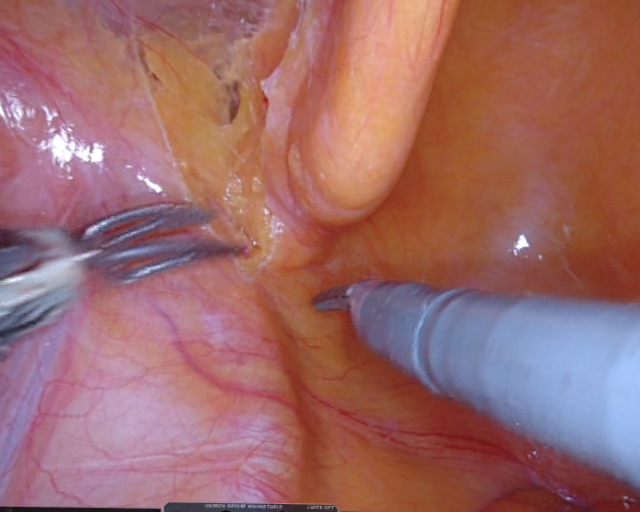}&
        \includegraphics[clip, angle=0, height=0.1\linewidth]{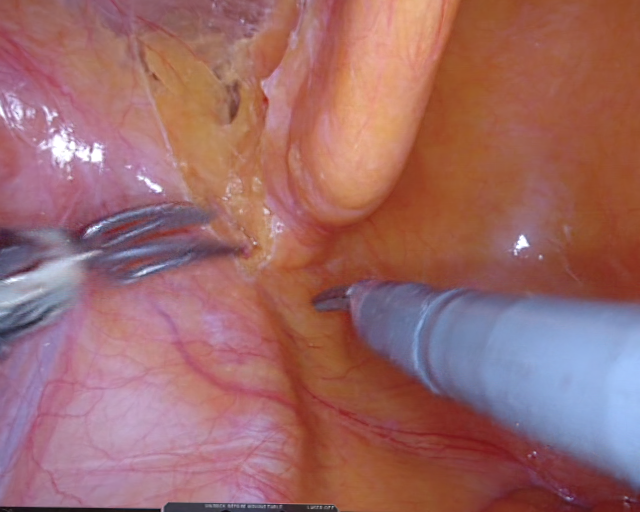}\\
        \includegraphics[clip, angle=0, height=0.1\linewidth]{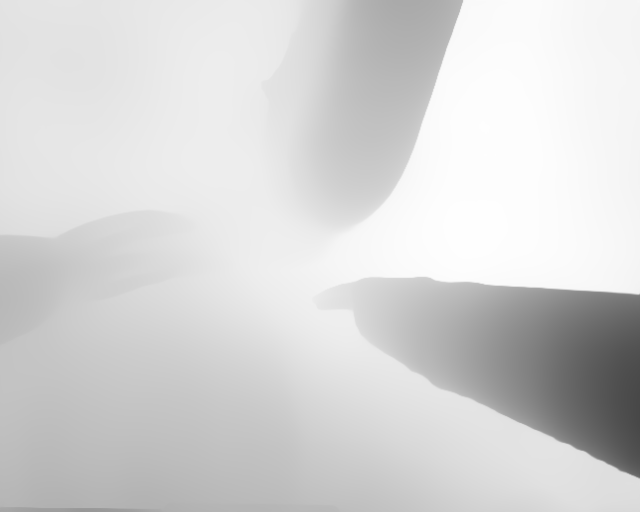}&
        \includegraphics[clip, angle=0, height=0.1\linewidth]{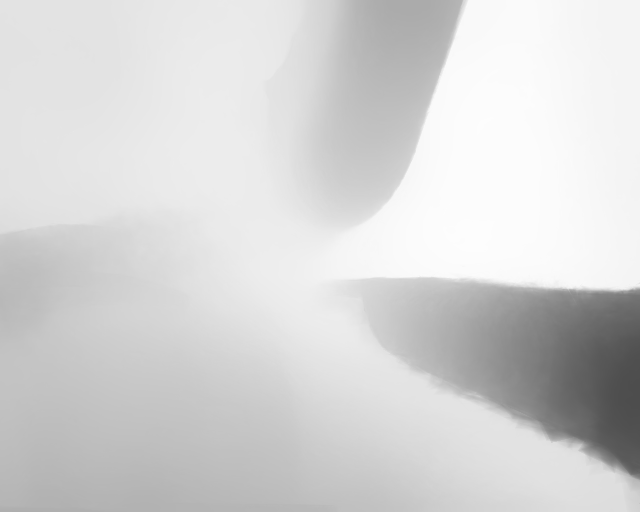}&
        \includegraphics[clip, angle=0, height=0.1\linewidth]{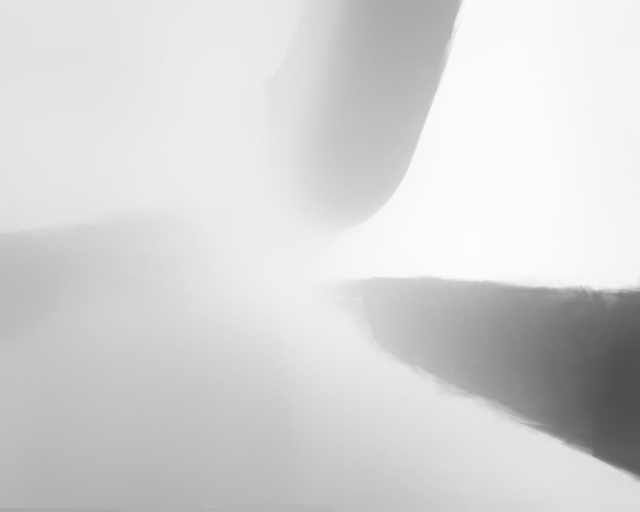}&
        \includegraphics[clip, angle=0, height=0.1\linewidth]{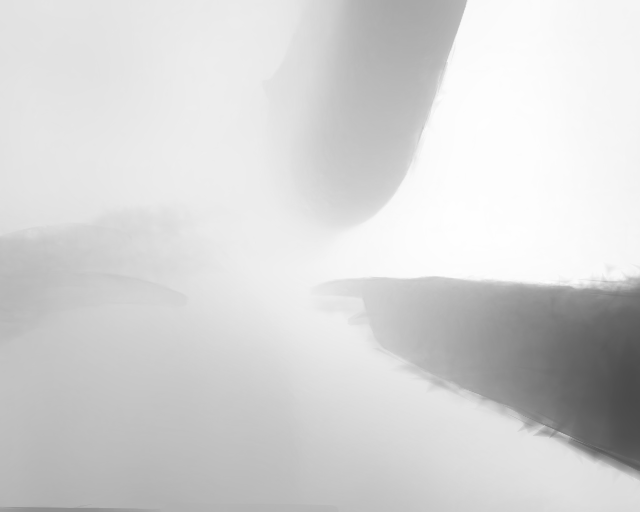}&
        \includegraphics[clip, angle=0, height=0.1\linewidth]{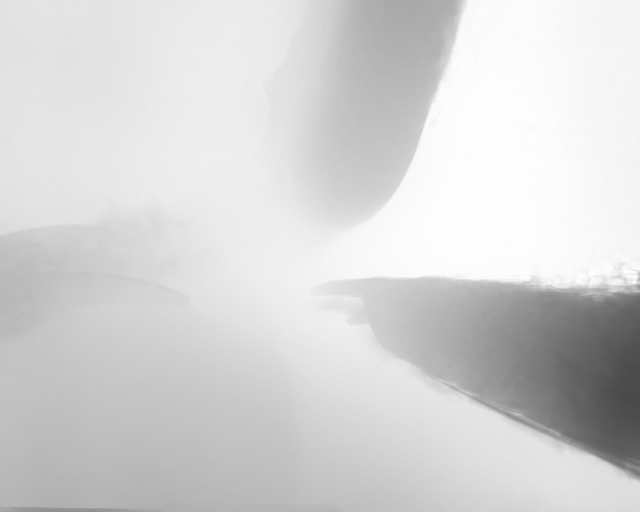}&
        \includegraphics[clip, angle=0, height=0.1\linewidth]{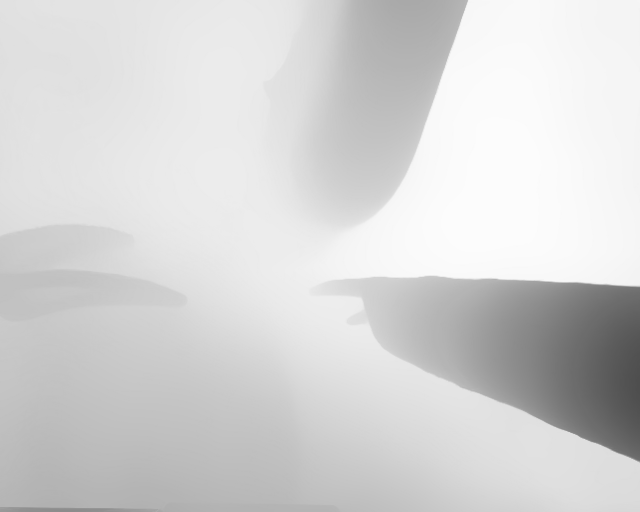}&
        \includegraphics[clip, angle=0, height=0.1\linewidth]{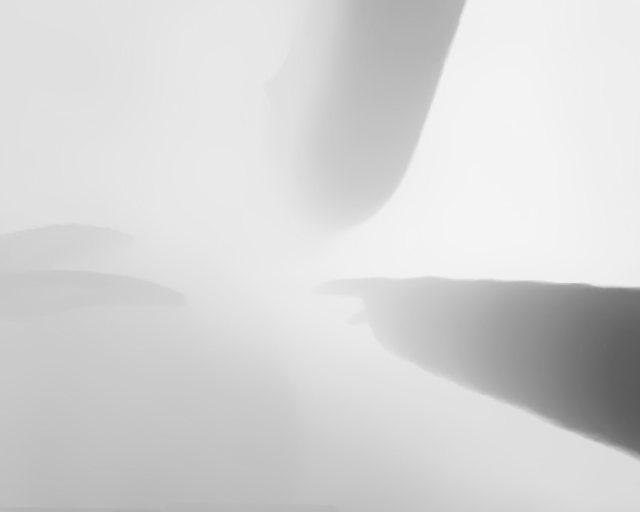} \\
        \includegraphics[clip, angle=0, height=0.1\linewidth]{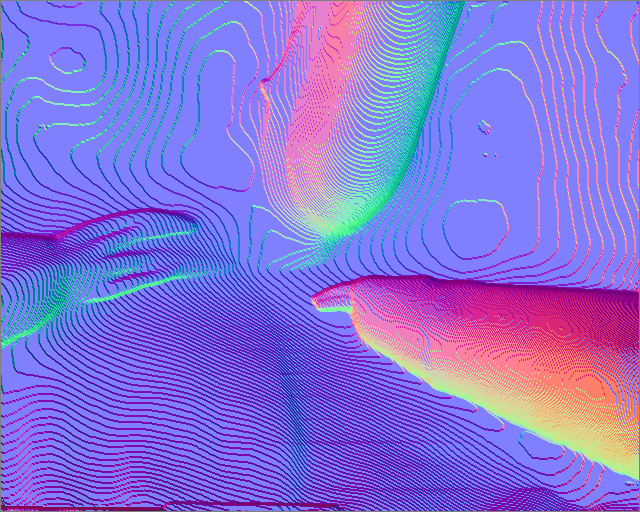}&
        \includegraphics[clip, angle=0, height=0.1\linewidth]{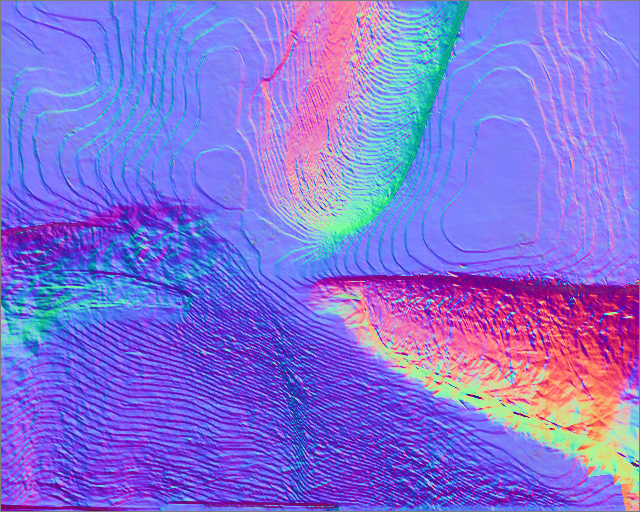}&
        \includegraphics[clip, angle=0, height=0.1\linewidth]{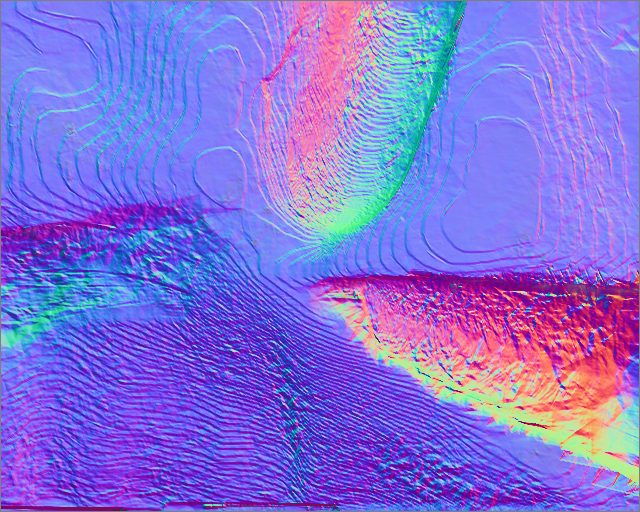}&
        \includegraphics[clip, angle=0, height=0.1\linewidth]{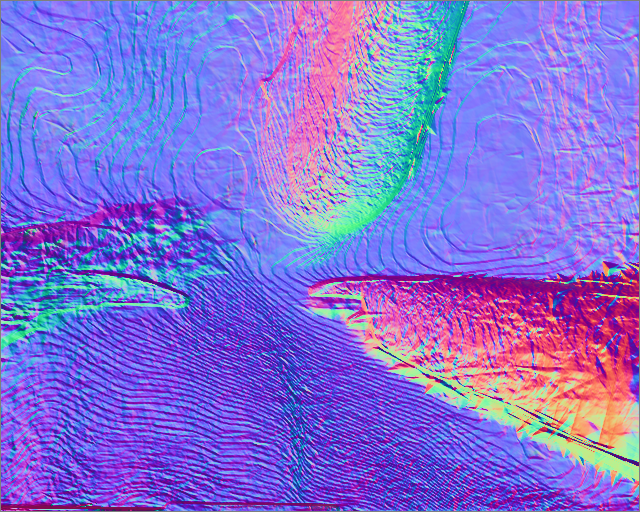}&
        \includegraphics[clip, angle=0, height=0.1\linewidth]{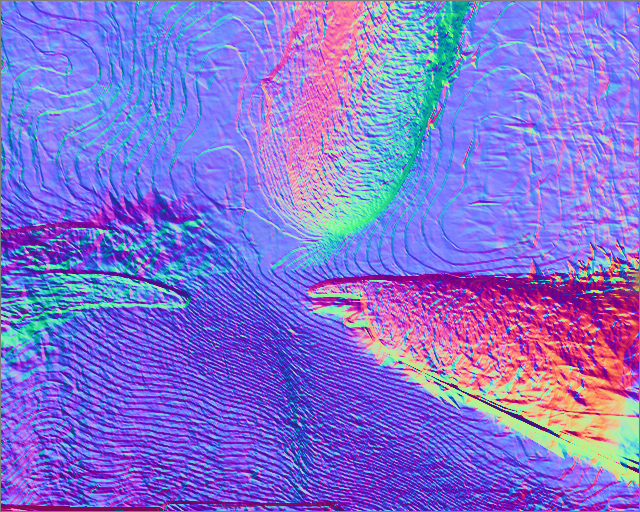}&
        \includegraphics[clip, angle=0, height=0.1\linewidth]{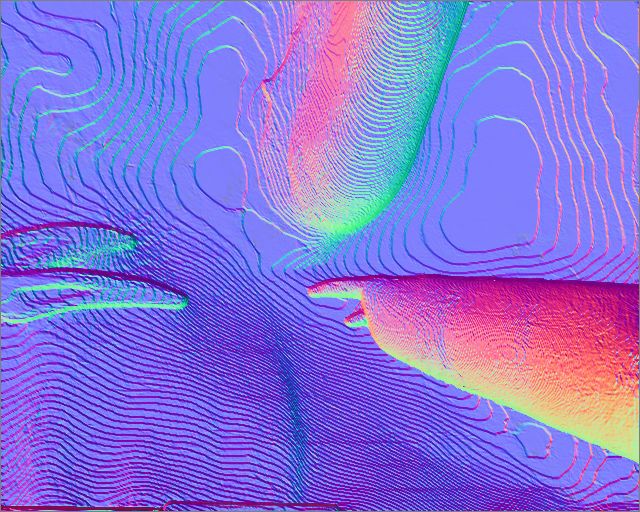}&
        \includegraphics[clip, angle=0, height=0.1\linewidth]{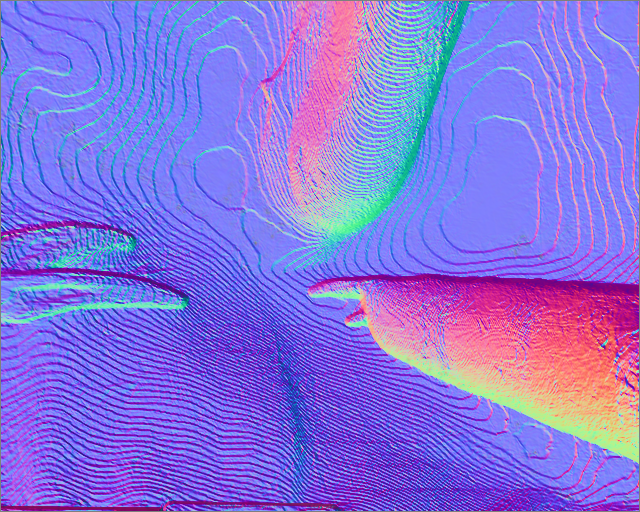}
    \end{tabular}
    }
    \caption{\textbf{Qualitative results of the ablation study} on a frame from the {\em pulling} sequence. 
    \textbf{Row 1:} RGB images. \textbf{Row 2:} Surface depth. \textbf{Row 3:} Surface normal.}
    \label{fig:ablation}
\end{figure}

First, we performed an ablation study on the key components of our loss function using part of the {\em pulling} sequence~\cite{endoNERF}. Starting from the full loss formulation in Eq.~\eqref{eq:total-loss}, we progressively removed the terms $\mathcal{L}_{\text{edge}}^{D}$, $\mathcal{L}_{\text{grad}}^{D}$, $\mathcal{L}_{\text{grad}}^{C}$, $\mathcal{L}_{\text{depth}}^{D}$ and $\mathcal{L}_{\text{depth}}^{C}$, one at a time. Quantitative results are reported in Table~\ref{tab:ablation}. We observe that removing the depth regularization terms substantially degrades geometric reconstruction quality, as the optimization then relies solely on the photometric loss terms. Consequently, we separate this ablation from the remaining ones with a dashed line and do not include it in the comparison of best-performing variants. Fig.~\ref{fig:ablation} provides a qualitative comparison illustrating these effects. Overall, the results reveal a  trade-off between the geometric accuracy and photometric consistency during optimization, supporting the inclusion of all loss terms in the final formulation.

We evaluated our SurgicalTS on the Endo-NeRF~\cite{endoNERF} dataset. Table~\ref{tab:endonerf} reports our results in comparison with other NeRF-, GS- or TS-based solutions~\cite{endogaussian,salort-benejam2026endodts,endoNERF}, providing competitive results in terms of photometric quality and consistently outperforming previous TS-based methods in terms of depth quality. These results demonstrate the superiority of our SurgicalTS in obtaining accurate and robust joint solutions, rather than focusing solely on capturing one of the properties. Consistent improvements can also be observed in Fig.~\ref{fig:endonerf}, where our method produces more geometrically coherent representations with smoother normals and fewer discontinuities or floating triangles than competing approaches. Nevertheless, in particularly challenging scenarios such as the {\em cutting} video, some local discontinuities still appear. The table also reports the training time ($T_t$) in minutes and the rendering speed (FPS) in testing on a single GPU NVIDIA RTX A5000. Thanks to our method a consistent reduction in training time is achieved compared to Endo-4DTS~\cite{salort-benejam2026endodts}, while the rendering speed is quite similar. Note that these values depend on the resolution of the input videos and the final number of triangles used to model the tissue. 

\begin{table}[h!]
\renewcommand{\arraystretch}{1.2}
\centering
\caption{\textbf{Quantitative comparison} of our SurgicalTS with EndoNeRF~\cite{endoNERF}, EndoGaussian~\cite{endogaussian}, and Endo-4DTS~\cite{salort-benejam2026endodts} on the EndoNeRF dataset.
Best results in \textbf{bold}.}\label{tab:endonerf}
\begin{tabular}{c|c|ccc|cc|cc}
 & {Methods} 
 & {PSNR}$\uparrow$ 
 & {SSIM}$\uparrow$ 
 & {LPIPS}$\downarrow$ 
 & {MAE}$\downarrow$ 
 & {RMSE}$\downarrow$ 
 & {FPS}$\uparrow$
 & ${T_{t}}\downarrow$\\
\hline

\multirow{4}{*}{\rotatebox{90}{Pulling}}
& EndoNeRF      & 34.217 & 0.938 & 0.160 & -- & --& --& --  \\
& EndoGaussian  & 37.308 & 0.958 & 0.070 & -- & --& --& -- \\
& Endo-4DTS     & \textbf{40.39} & \textbf{0.971} & \textbf{0.026} & 7.171 & 16.803 &\textbf{16.50}&329\\
& SurgicalTS (Ours) 
                 & \underline{39.890}&\underline{0.969}& \underline{0.027}& \textbf{2.259}& \textbf{6.539}&14.95 & \textbf{221}\\
\hline

\multirow{4}{*}{\rotatebox{90}{Cutting}}
& EndoNeRF      & 34.186 & 0.932 & 0.151 & -- & -- & --& --  \\
& EndoGaussian  & \underline{38.287} & \underline{0.962} & 0.058 & -- &  --& --& --  \\
& Endo-4DTS     & \textbf{38.876} & \textbf{0.966} & \textbf{0.029} & 5.759 & 15.593&\textbf{18.98}& 353\\
& SurgicalTS (Ours) 
                 & 36.468&0.954& \underline{0.045}& \textbf{2.715}& \textbf{8.018}& 17.25 & \textbf{242}

 \\
\hline

\end{tabular}
\end{table}

\newcolumntype{?}[1]{!{\vrule width #1}}
\begin{figure}[h!]
    \centering    
    \begin{tabular}{@{}ccc@{\hspace{1.5mm}}?{0.5mm}@{\hspace{1.5mm}}ccc@{}}
        GT& Endo-4DTS&Ours& GT&Endo-4DTS&Ours\\
        
        \includegraphics[clip, angle=0, height=0.12\linewidth]{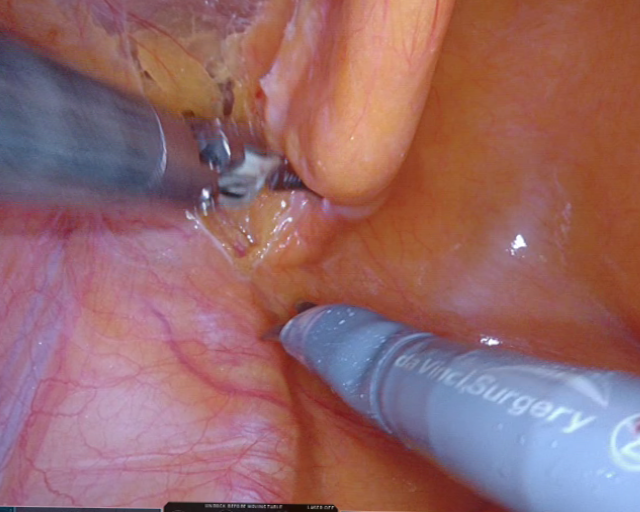}&
        \includegraphics[clip, angle=0, height=0.12\linewidth]{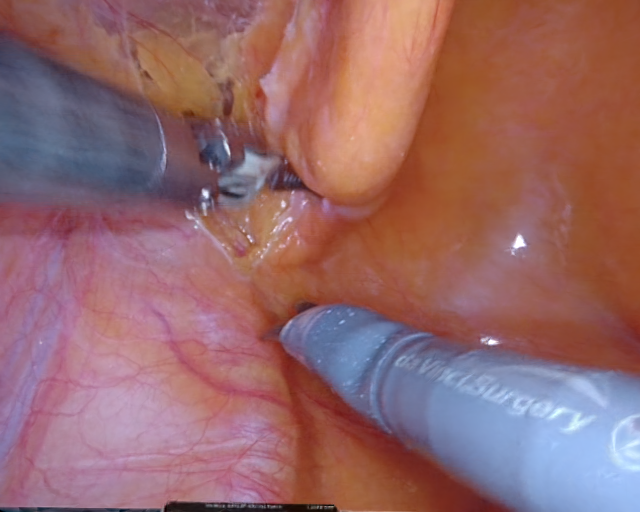}&
        \includegraphics[clip, angle=0, height=0.12\linewidth]{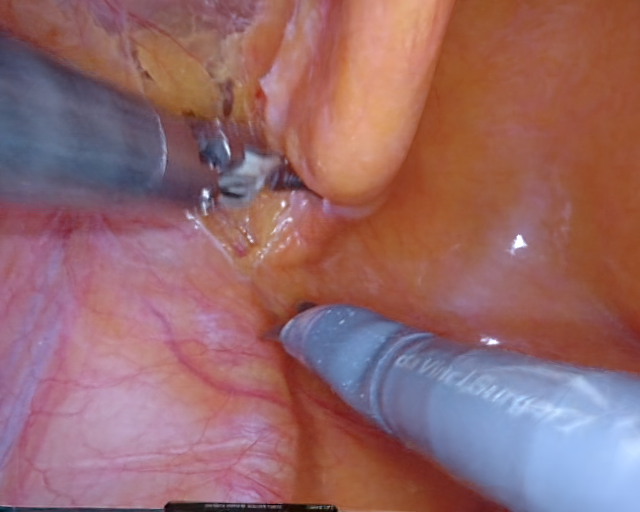}&
        \includegraphics[clip, angle=0, height=0.12\linewidth]{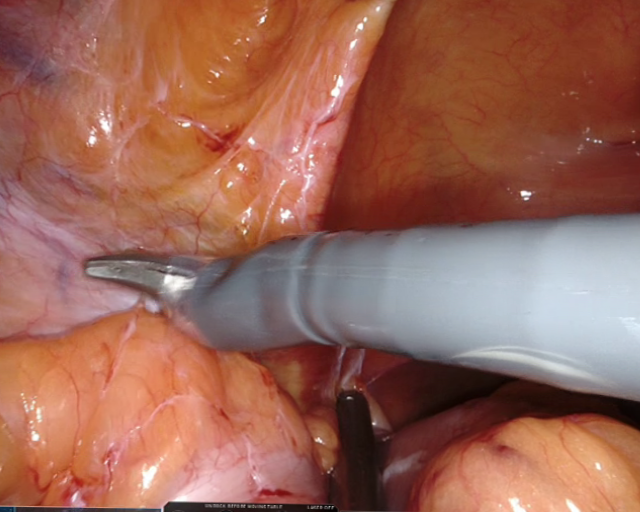}&
        \includegraphics[clip, angle=0, height=0.12\linewidth]{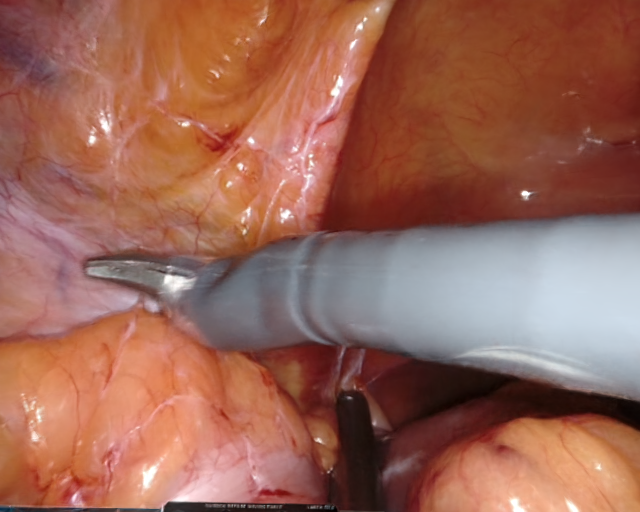}&
        \includegraphics[clip, angle=0, height=0.12\linewidth]{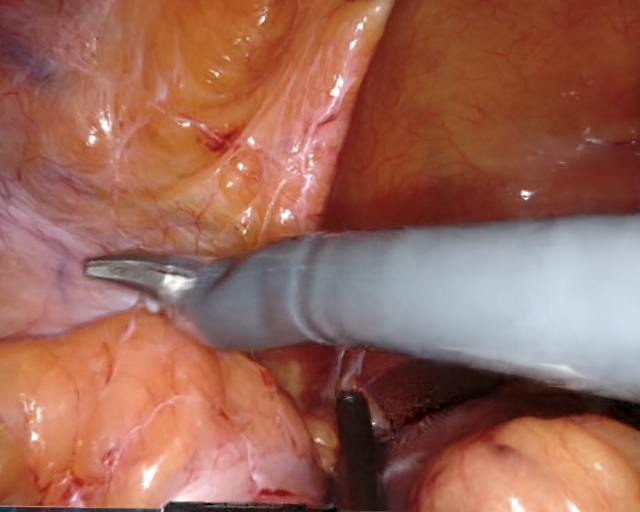}
        \\
         
        \includegraphics[clip, angle=0, height=0.12\linewidth]{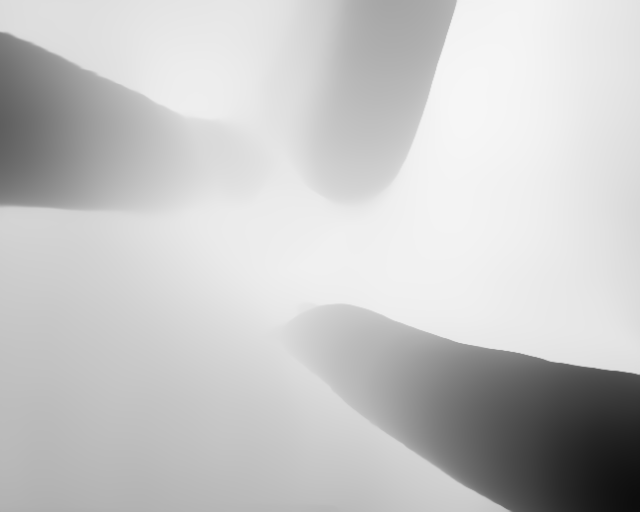}&
        \includegraphics[clip, angle=0, height=0.12\linewidth]{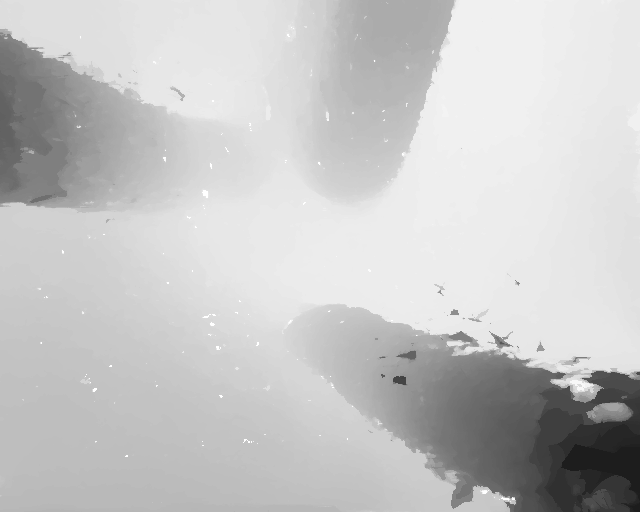}&
        \includegraphics[clip, angle=0, height=0.12\linewidth]{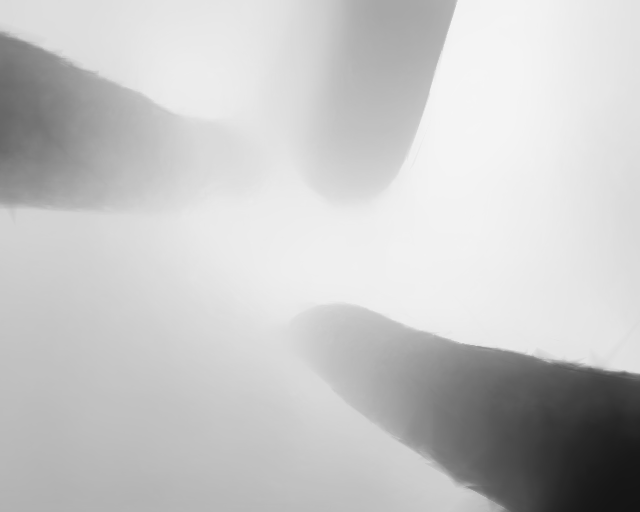}&
        \includegraphics[clip, angle=0, height=0.12\linewidth]{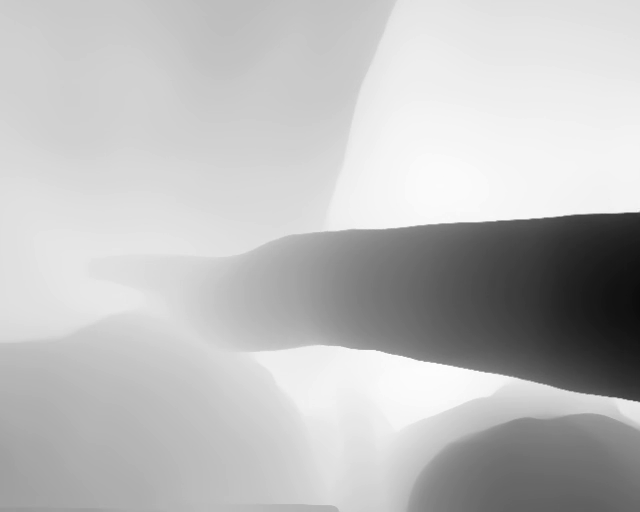}&
        \includegraphics[clip, angle=0, height=0.12\linewidth]{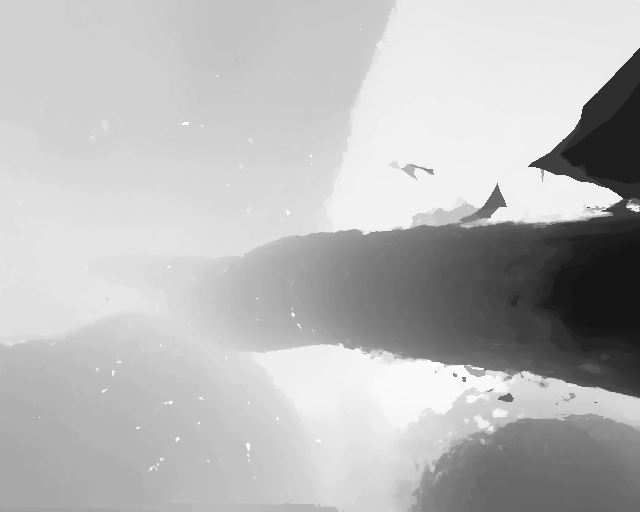}&
        \includegraphics[clip, angle=0, height=0.12\linewidth]{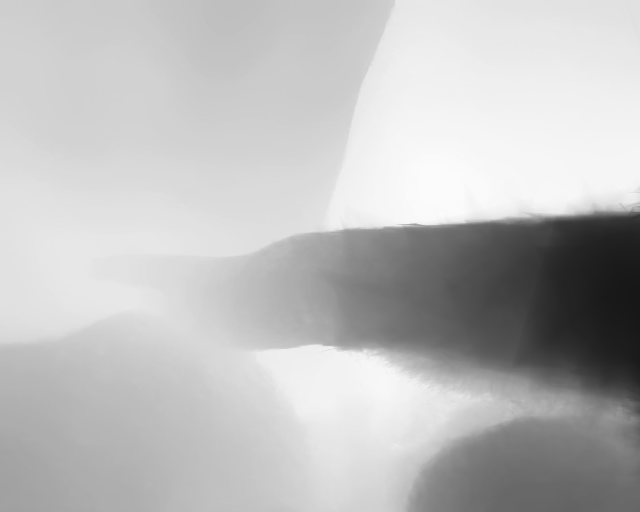}    
         \\
         
        \includegraphics[clip, angle=0, height=0.12\linewidth]{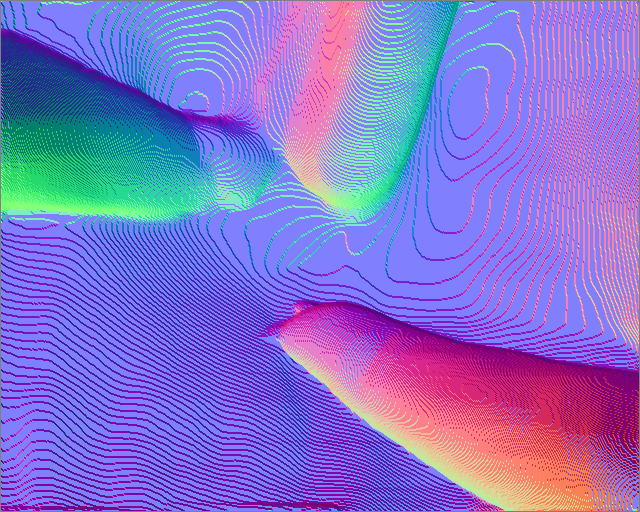}&
        \includegraphics[clip, angle=0, height=0.12\linewidth]{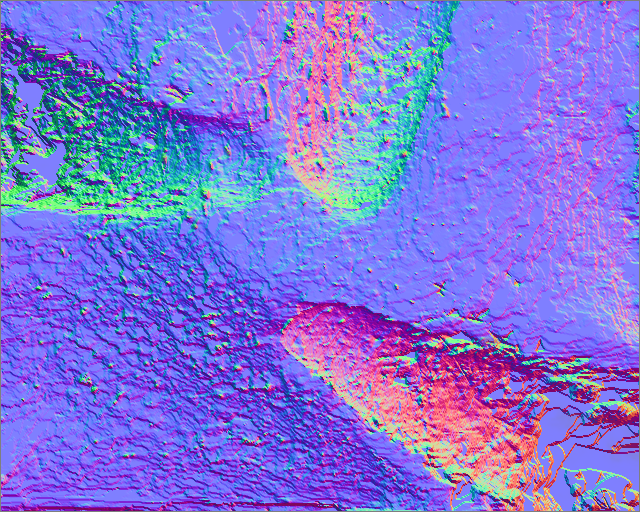}&
        \includegraphics[clip, angle=0, height=0.12\linewidth]{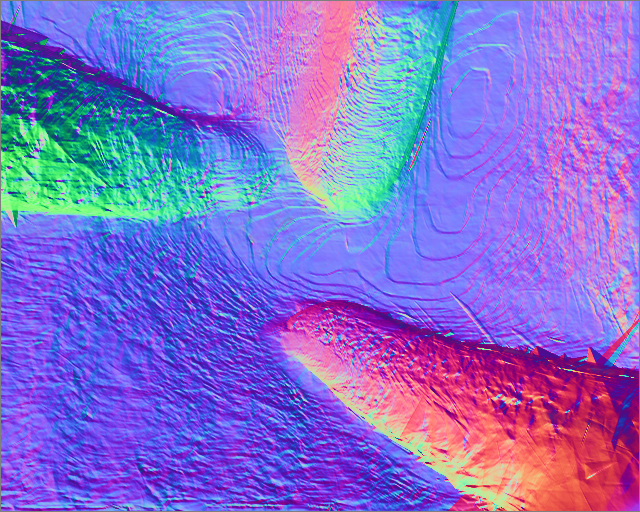}&
        \includegraphics[clip, angle=0, height=0.12\linewidth]{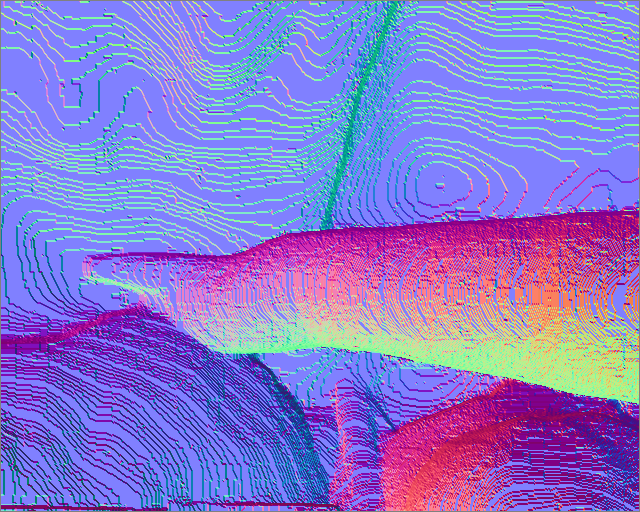}&
        \includegraphics[clip, angle=0, height=0.12\linewidth]{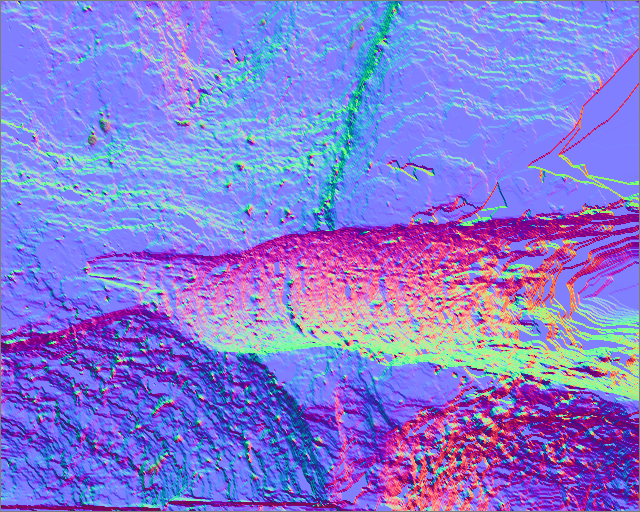}&
        \includegraphics[clip, angle=0, height=0.12\linewidth]{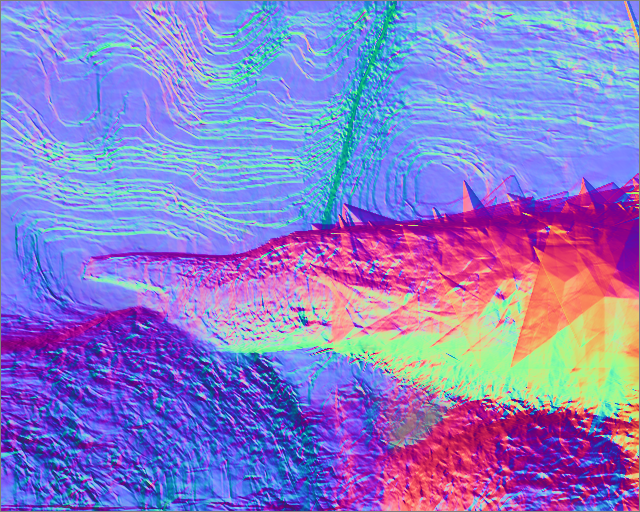}
          \\
    \end{tabular}
       \caption{\textbf{Qualitative evaluation} on EndoNeRF dataset~\cite{endoNERF}. On both sides rows show from top to bottom: RGB, Depth, and Normals images. \textbf{Left:} Pulling. \textbf{Right:} Cutting.}
       \label{fig:endonerf}
\end{figure}

\begin{figure}[h!]
    \centering
    \setlength{\tabcolsep}{2pt}
    \begin{tabular}[C]{@{}cccc?{0.5mm}cccc@{}}
        \begin{minipage}[c]{0.11\textwidth}
        \vspace{0.1cm}
          \includegraphics[width=\textwidth]{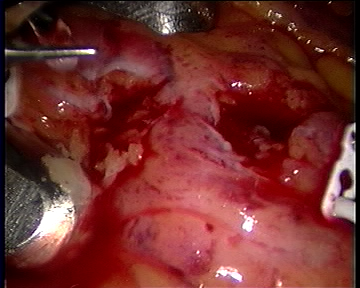}
        \end{minipage}&
        \begin{minipage}[c]{0.11\textwidth}
        \vspace{0.1cm}
          \includegraphics[width=\textwidth]{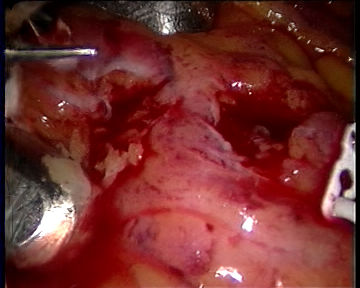}
        \end{minipage}& 
        \begin{minipage}[c]{0.11\textwidth}
        \vspace{0.1cm}
          \includegraphics[ width=\textwidth]{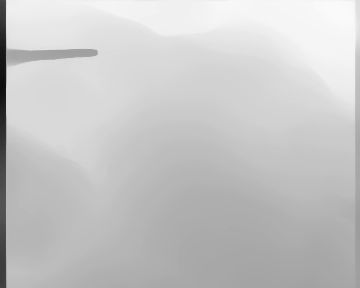}
        \end{minipage}&
        \begin{minipage}[c]{0.11\textwidth}
        \vspace{0.1cm}
          \includegraphics[width=\textwidth]{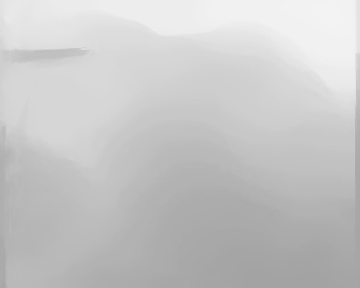}
        \end{minipage}&
        \begin{minipage}[c]{0.11\textwidth}
        \vspace{0.1cm}
          \includegraphics[width=\textwidth]{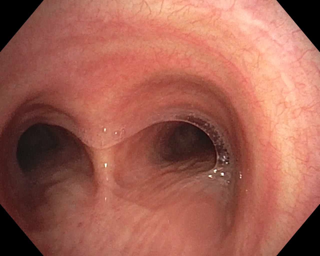}
        \end{minipage}&
        \begin{minipage}[c]{0.11\textwidth}
        \vspace{0.1cm}
          \includegraphics[width=\textwidth]{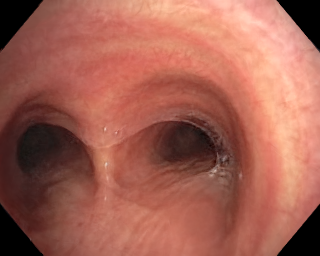}
        \end{minipage}& 
        \begin{minipage}[c]{0.11\textwidth}
        \vspace{0.1cm}
          \includegraphics[ width=\textwidth]{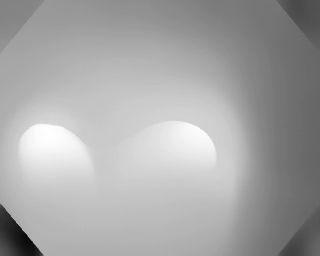}
        \end{minipage}&
        \begin{minipage}[c]{0.11\textwidth}
        \vspace{0.1cm}
          \includegraphics[width=\textwidth]{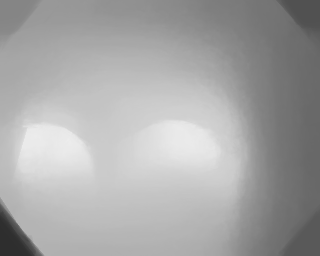}
        \end{minipage}  \\

        \begin{minipage}[c]{0.11\textwidth}
        \vspace{0.1cm}
          \includegraphics[width=\textwidth]{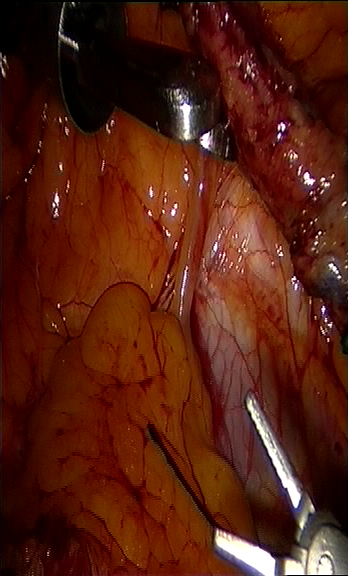}
        \end{minipage}&
        \begin{minipage}[c]{0.11\textwidth}
        \vspace{0.1cm}
          \includegraphics[width=\textwidth]{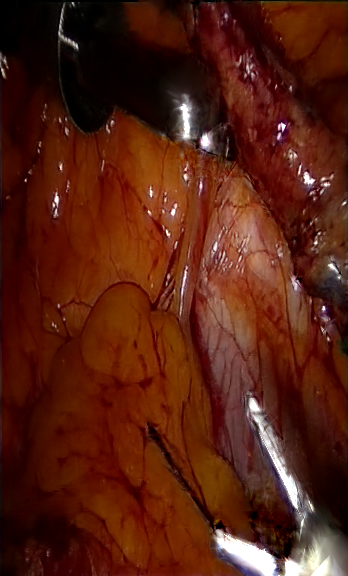}
        \end{minipage}& 
        \begin{minipage}[c]{0.11\textwidth}
        \vspace{0.1cm}
          \includegraphics[ width=\textwidth]{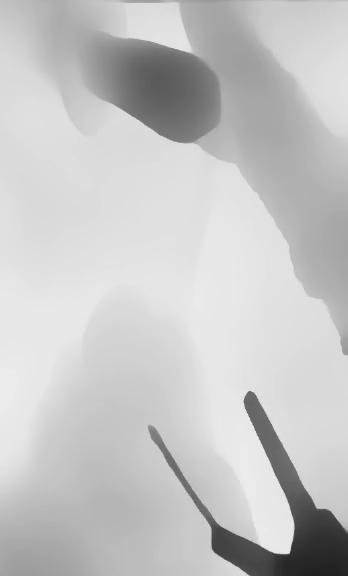}
        \end{minipage}&
        \begin{minipage}[c]{0.11\textwidth}
        \vspace{0.1cm}
          \includegraphics[width=\textwidth]{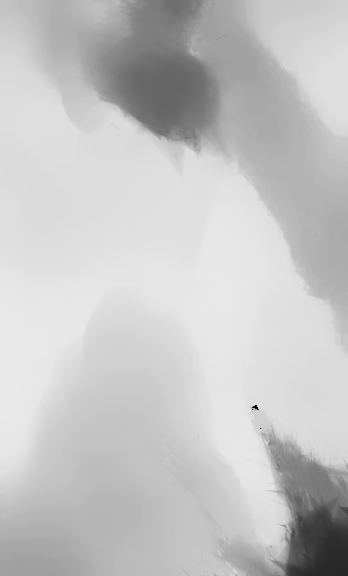}
        \end{minipage}&
        \begin{minipage}[c]{0.11\textwidth}
        \vspace{0.1cm}
        \includegraphics[width=\textwidth]{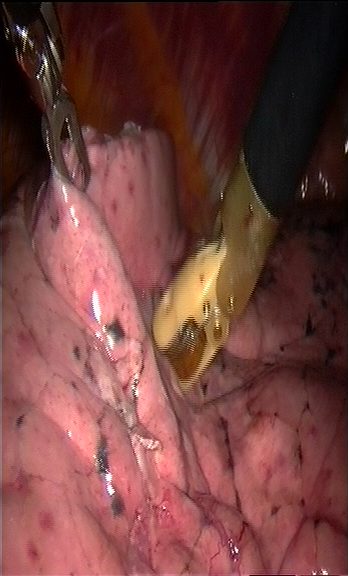}
        \end{minipage}&
        \begin{minipage}[c]{0.11\textwidth}
        \vspace{0.1cm}
          \includegraphics[width=\textwidth]{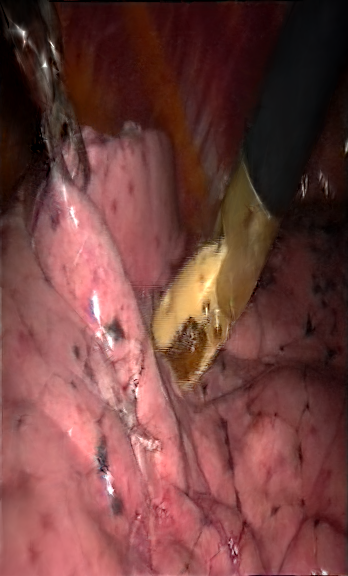}
        \end{minipage}& 
        \begin{minipage}[c]{0.11\textwidth}
        \vspace{0.1cm}
          \includegraphics[ width=\textwidth]{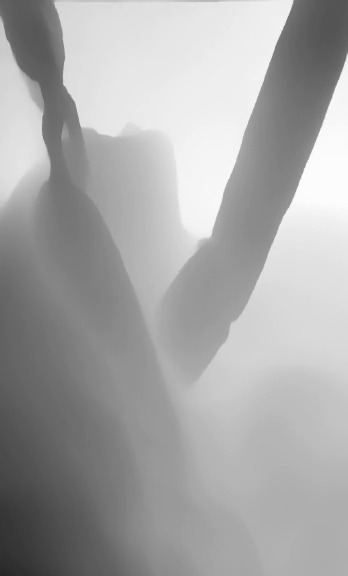}
        \end{minipage}&
        \begin{minipage}[c]{0.11\textwidth}
        \vspace{0.1cm}
          \includegraphics[width=\textwidth]{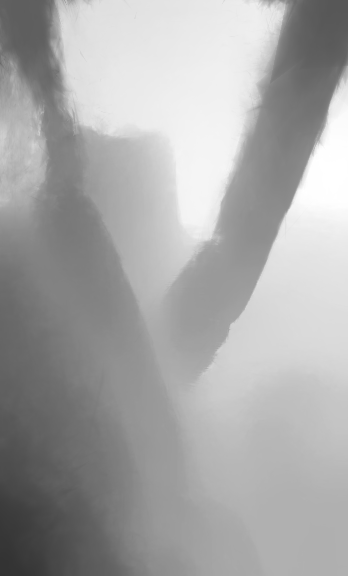 }
        \end{minipage} \\
    \end{tabular} 
     \caption{\textbf{Qualitative evaluation} on NeRFscopy dataset~\cite{salortISBI26}. On both sides columns show from left to right: input and rendered RGB, input and rendered depth. \textbf{Left:} TECAB1 and TECAB2 images. \textbf{Right:} Bronchoscopy and lung lobectomy images.}
     \label{fig:nerfscopy}
     
\end{figure}

We also report results on the NeRFscopy dataset~\cite{salortISBI26}. These results are provided in Table~\ref{tab:nerfscopy}, demonstrating superior performance over previous methods~\cite{salort-benejam2026endodts,salortISBI26,endoNERF} in both photometric and depth qualities, and showing the effectiveness of our model in capturing various tissues. In terms of efficiency, our method shows competitive solutions for medical scenarios that slightly exceeds the performance of Endo-4DTS~\cite{salort-benejam2026endodts} and clearly outperforms NeRF-based solutions that rarely reach 1 FPS~\cite{salortISBI26,endoNERF}. A qualitative evaluation of these results is displayed in Fig.~\ref{fig:nerfscopy}, observing as our joint estimation is physically possible.
        
\begin{table}[h!]
\renewcommand{\arraystretch}{1.2}

\centering
\caption{\textbf{Quantitative comparison} of our SurgicalTS with EndoNeRF~\cite{endoNERF}, NeRFscopy~\cite{salortISBI26}, and Endo-4DTS~\cite{salort-benejam2026endodts} on the NeRFscopy dataset~\cite{salortISBI26}.
Best results in \textbf{bold}.}\label{tab:nerfscopy}
\begin{tabular}{c|c|ccc|cc|cc}
 & {Methods} 
 & {PSNR}$\uparrow$ 
 & {SSIM}$\uparrow$ 
 & {LPIPS}$\downarrow$ 
 & {MAE}$\downarrow$ 
 & {RMSE}$\downarrow$
 & {FPS}$\uparrow$
 & ${T_{t}}\downarrow$\\
\hline

\multirow{4}{*}{\rotatebox{90}{TECAB1}}
& EndoNeRF      & 25.791&  0.742& 0.255 & -- & --  & -- & --\\
& NeRFscopy  &25.811& 0.750& 0.255 & -- & --& -- & --\\
& Endo-4DTS     &  \underline{28.260}&  \underline{0.863}& \underline{0.136} & 4.762& 15.358  & 41.30& 158\\
& SurgicalTS (Ours) 
                 & \textbf{29.473}&\textbf{0.894}&	\textbf{0.096} & \textbf{4.721}&\textbf{14.805} &\textbf{48.88}& \textbf{80}
\\
\hline
\multirow{4}{*}{\rotatebox{90}{TECAB2}}
& EndoNeRF      & 24.954&  0.685& 0.419 & -- & -- & -- & -- \\
& NeRFscopy  & 24.864&0.689& 0.429 & -- & -- & --  & --\\
& Endo-4DTS     & \textbf{25.983}& \textbf{ 0.783} &\textbf{0.238} & 9.525& 18.138  &20.18& 194\\
& SurgicalTS (Ours) 
                 &  \underline{25.009}& \underline{0.765}&  \underline{0.247}& \textbf{6.003}& \textbf{13.817}&\textbf{20.45}& \textbf{144}\\
\hline

\multirow{4}{*}{\rotatebox{90}{Lung lob.}}
& EndoNeRF      &27.142&  0.788& 0.293 & -- & -- & -- & -- \\
& NeRFscopy  &  27.285&  0.791& 0.275& -- & --  & -- & --\\
& Endo-4DTS     & \underline{28.435}&\underline{0.825}& \textbf{0.178} &  7.895& 18.759 &21.75& 191\\
& SurgicalTS (Ours) 
                 &\textbf{28.673}&\textbf{0.833}& \underline{0.189} & \textbf{5.027} & \textbf{12.124} & \textbf{22.17}& \textbf{139}\\
\hline

\multirow{4}{*}{\rotatebox{90}{Bronco}}
& EndoNeRF      & 33.872&  0.867& 0.588 & -- & -- & -- & -- \\
& NeRFscopy  &  34.405& 0.875& 0.156 & -- & --  & -- & --\\
& Endo-4DTS     & \underline{35.566} & \underline{0.902} & \underline{0.080} & 2.581
& 6.790 &24.22& 125\\
& SurgicalTS (Ours) 
                 &\textbf{36.304}  &\textbf{0.914}  & \textbf{0.074} & \textbf{2.051} &\textbf{ 5.731} & \textbf{25.12}& \textbf{62}\\
\hline
\end{tabular}
\end{table}

\section{Conclusion}
We presented SurgicalTS, a novel triangle-splatting framework for self-supervised 4D reconstruction of deformable endoscopy scenes. By introducing a deformation-aware initialization, re-parametrizing the variables to be estimated in the deformation network, and using spatio-temporal physical regularizations, our method improves geometric fidelity while maintaining competitive photometric quality and rendering speed. Extensive experiments on in-vivo datasets show consistent improvements, demonstrating our superior performance over state-of-the-art methods. These results emphasize the importance of including physically-aware regularizations during optimization, as well as a good initialization for robust 4D reconstruction of endoscopy scenes.

\begin{credits}
\subsubsection{\ackname} This work has been supported by the project GRAVATAR PID2023-151184OB-I00 funded by MCIU/AEI/10.13039/501100011033 and by ERDF, UE; and by the Government of Catalonia under 2025 FI-STEP 00398.

\subsubsection{\discintname}  The authors have no competing interests to declare that are relevant to the content of this article.

\end{credits}

\bibliographystyle{splncs04}
\bibliography{references}
\end{document}